\documentclass[letterpaper]{article} 
\usepackage[]{aaai24}
\usepackage{times}  
\usepackage{helvet}  
\usepackage{courier}  
\usepackage[hyphens]{url}  
\usepackage{graphicx} 
\usepackage{natbib}  
\usepackage{caption} 
\usepackage{algorithm}
\usepackage{algorithmic}

\usepackage{arydshln}
\usepackage{multirow}
\usepackage{multicol}
\usepackage{amsmath}
\usepackage{bm}
\usepackage{url}
\usepackage{underscore}
\usepackage{amsfonts}
\usepackage{xcolor}
\usepackage{color, colortbl}
\usepackage{makecell}
\usepackage{enumitem}
\usepackage{colortbl}
\usepackage{booktabs}
\usepackage{pifont}
\usepackage{hyperref}
\usepackage[switch]{lineno}

\newcommand{\hyf}[1]{{\color{black}#1}}

\newcommand{\wen}[1]{{\color{black}#1}}
\newcommand{\tjj}[1]{{\color{black}#1}}

\usepackage{newfloat}
\usepackage{listings}
\DeclareCaptionStyle{ruled}{labelfont=normalfont,labelsep=colon,strut=off} 
\floatstyle{ruled}
\newfloat{listing}{tb}{lst}{}
\floatname{listing}{Listing}
\title{Structure-CLIP: Towards Scene Graph 
Knowledge to Enhance Multi-modal Structured Representations}
\author{
    Yufeng Huang \textsuperscript{\rm 1} \equalcontrib, 
    Jiji Tang \textsuperscript{\rm 3} \equalcontrib, 
    Zhuo Chen \textsuperscript{\rm 2}, 
    Rongsheng Zhang \textsuperscript{\rm 2,3},
    Xinfeng Zhang \textsuperscript{\rm 3},
    Weijie Chen \textsuperscript{\rm 3},
    Zeng Zhao \textsuperscript{\rm 3},
    Zhou Zhao \textsuperscript{\rm 2}, 
    Tangjie Lv,\textsuperscript{\rm 3},
    Zhipeng Hu \textsuperscript{\rm 3},
    Wen Zhang \textsuperscript{\rm 1}\thanks{Corresponding Author.}}
\affiliations{
    \textsuperscript{\rm 1}School of Software Technology, Zhejiang University\\
    \textsuperscript{\rm 2}College of Computer Science and Technology, Zhejiang University\\
    \textsuperscript{\rm 3}Fuxi AI Lab, Netease Inc. \\
    \{huangyufeng, zhuo.chen, zhaozhou, zhang.wen\}@zju.edu.cn \\
    \{tangjiji01, zhangrongsheng, zhangxinfeng01, chenweijie05, hzlvtangjie, zphu\}@corp.netease.com  \\
}

\usepackage{bibentry}

\begin{document}

\maketitle

\begin{abstract}
Large-scale vision-language pre-training has achieved significant performance in multi-modal understanding and generation tasks.
However, existing methods often perform poorly on image-text matching tasks that require structured representation\hyf{s}, i.e., representations of objects, attributes, and relations. \hyf{As illustrated in Fig.~\ref{fig:case} (a),} the models cannot make a \tjj{distinction} between ``An astronaut rides a horse" and ``A horse rides an astronaut". 
This is because they fail to fully leverage structured knowledge when learning representations in multi-modal scenarios.
In this paper, we present an end-to-end framework Structure-CLIP, which integrates \tjj{\emph{Scene Graph Knowledge} (SGK)} to enhance multi-modal structured representations. 
Firstly, we use scene graphs to guide the \tjj{construction} of \emph{semantic negative} examples, which results in an increased emphasis on learning structured representations.
\tjj{Moreover, a \emph{Knowledge-Enhance Encoder} (KEE) is proposed to leverage 
\hyf{SGK} as input to further enhance structured representations.}
To verify the effectiveness of the proposed framework, we \tjj{pre-train} our model with the aforementioned approaches and conduct experiments on 
downstream tasks. 
\tjj{Experimental results demonstrate that Structure-CLIP achieves \emph{state-of-the-art} (SOTA) performance on VG-Attribution and VG-Relation datasets, 
with $12.5\%$ 
 and $4.1\%$  ahead of the multi-modal SOTA model respectively.}
Meanwhile, the results on MSCOCO indicate that Structure-CLIP significantly enhances the structured representations while maintaining the ability of general representations.
Our code is available at \textcolor{blue}{\url{https://github.com/zjukg/Structure-CLIP}}.
\end{abstract}

\section{Introduction}

\begin{figure}[t]
  \centering
  \includegraphics[width = 1\linewidth]{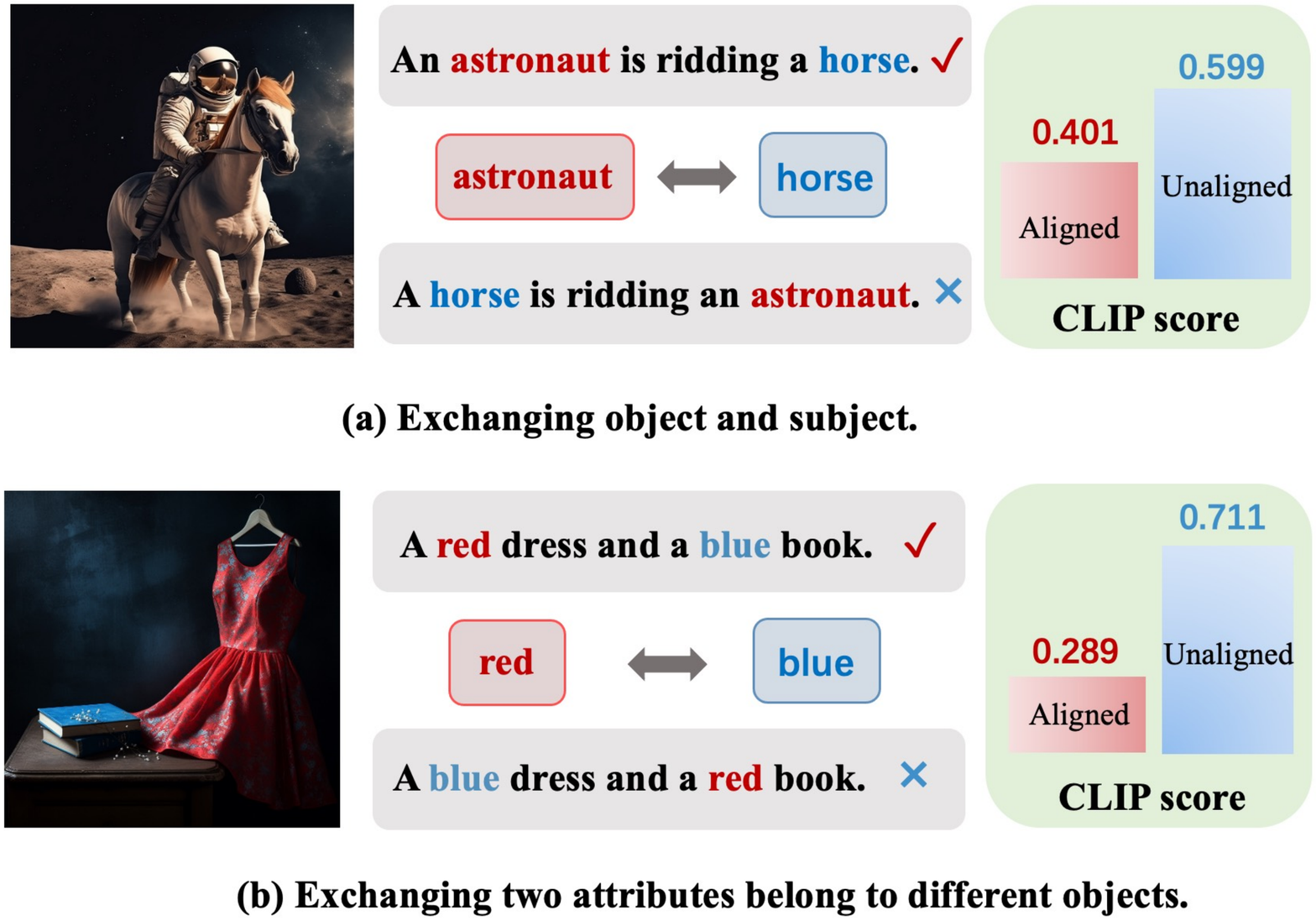}

  \caption{CLIP scores \tjj{(after normalizing among two results)} between the image and aligned/unaligned captions. The results show that the CLIP model does not have the ability to distinguish sentences with structured semantic differences. }
  \label{fig:case}
  \vspace{-10pt}
\end{figure}

\tjj{\emph{Vision-language models} (VLMs)} have demonstrated significant performance in various multi-modal understanding and generation tasks \cite{DBLP:conf/icml/RadfordKHRGASAM21, DBLP:conf/icml/0001LXH22,DBLP:conf/cvpr/SinghHGCGRK22,DBLP:journals/corr/abs-1908-03557}.
Despite the impressive performance of multi-modal models in various tasks, the question of whether these models can effectively capture structured knowledge(i.e., the ability to comprehend object properties and the relationships between objects) remains unresolved.

For example, as shown in \tjj{Fig. \ref{fig:case}} (a), the CLIP score (i.e., semantic similarity) between the image and the correctly matched caption (``An astronaut is riding a horse"), exhibits a lower value in contrast to the score between the image and a non-matching caption (``A horse is riding an astronaut"). Subsequently, \tjj{Fig. \ref{fig:case}} (b) illustrate\wen{s} that exchanging attributes between two objects can also pose challenges for the model to accurately distinguish 
\wen{their semantics.}
 These findings suggest that the generic representations yielded by the CLIP model are unable to differentiate between text segments that encompass identical words but diverge in terms of structured knowledge.
In other words, the CLIP model exhibits a tendency similar to that of a bag-of-words approach, which does not understand fine-grained semantics in sentences ~\cite{lin2023visualgptscore}.

Winoground ~\cite{DBLP:conf/cvpr/ThrushJBSWKR22} 
\wen{is the first work focusing}
on this problem and performed a broad-based examination. They intentionally created a dataset consisting of 400 instances, where each instance consists of two sentences with identical word compositions but different semantic meanings.
They evaluated various well-performing VLMs (e.g., VinVL ~\cite{DBLP:conf/cvpr/ZhangLHY0WCG21}, UNITER ~\cite{DBLP:conf/eccv/ChenLYK0G0020},  ViLBERT ~\cite{DBLP:conf/nips/LuBPL19}, and CLIP ~\cite{DBLP:conf/icml/RadfordKHRGASAM21}), with the aim of assessing the structured representations pertaining to objects, attributes, and relations. Unfortunately, their findings indicate that the outcomes are on par with a random selection, despite these models demonstrating human-level proficiency in other tasks.
The results of these tasks demonstrate that general representations are insufficient 
\wen{for}
semantic comprehension. 
It is thus inferred that an increased emphasis should be placed on structured representations.

NegCLIP \cite{DBLP:journals/corr/abs-2210-01936} enhances structured representations by integrating task-specific negative samples, which are generated by randomly exchanging any two words in a sentence. Thus, while general representations maintain consistency in positive and negative samples, structured representations exhibit divergence. Employing the contrastive learning approach, it compels the model to acquire structured representations rather than general representations.
Moreover, NegCLIP also provides a large-scale test bed to evaluate the \hyf{capabilities} of VLMs in terms of structured representations.
Nevertheless, NegCLIP suffers from a lack of understanding and modeling of the semantic knowledge during negative sample construction, which results in a notable deterioration in the quality of negative examples.
For example, when the attributes ``white" and ``black" are interchanged in the original caption ``Black and white cows", the underlying semantic meaning of the sentence remains invariant. 
 Such low-quality negative examples further lead to performance degradation.

In this paper, we propose Structure-CLIP, a novel approach that leverages \tjj{\emph{Scene Graph Knowledge} (SGK)} to enhance multi-modal structured representations. Firstly, in contrast to the random swap method in NegCLIP, we utilize SGK to construct word swaps that better match the underlying intent.
Secondly,  we propose a \tjj{\emph{Knowledge-Enhanced Encoder} (KEE)}, leveraging SGK to extract essential structure information. 
By incorporating structured knowledge at the input level, \tjj{the proposed KEE} can further enhance the \tjj{ability} of structured representations.
Results on Visual Genome Relation and Visual Genome Attribution show the \tjj{\emph{state-of-the-art} (SOTA)} performance of Structure-CLIP and the effectiveness of its components.
Additionally, we perform cross-modal retrieval evaluations on MSCOCO, which shows that Structure-CLIP still retains sufficient general representation ability.

Overall, our contributions are three-fold:
\begin{itemize}[leftmargin=*]
\item To the best of our knowledge, Structure-CLIP is the first method to enhance detailed structured representations by constructing \emph{Semantic Negative} samples.
\item A \emph{Knowledge-Enhanced Encoder} is introduced in Structure-CLIP to leverage the structured knowledge as the input to enhance structured representations.
\item 
\hyf{We conduct comprehensive experiments demonstrating that Structure-CLIP is able to achieve SOTA performance on \tjj{structured representations downstream} tasks and yield significant improvements \tjj{on} structured representations.} 
\end{itemize}

\begin{figure*}[!htbp]
  \centering
  \includegraphics[width=0.95\linewidth]{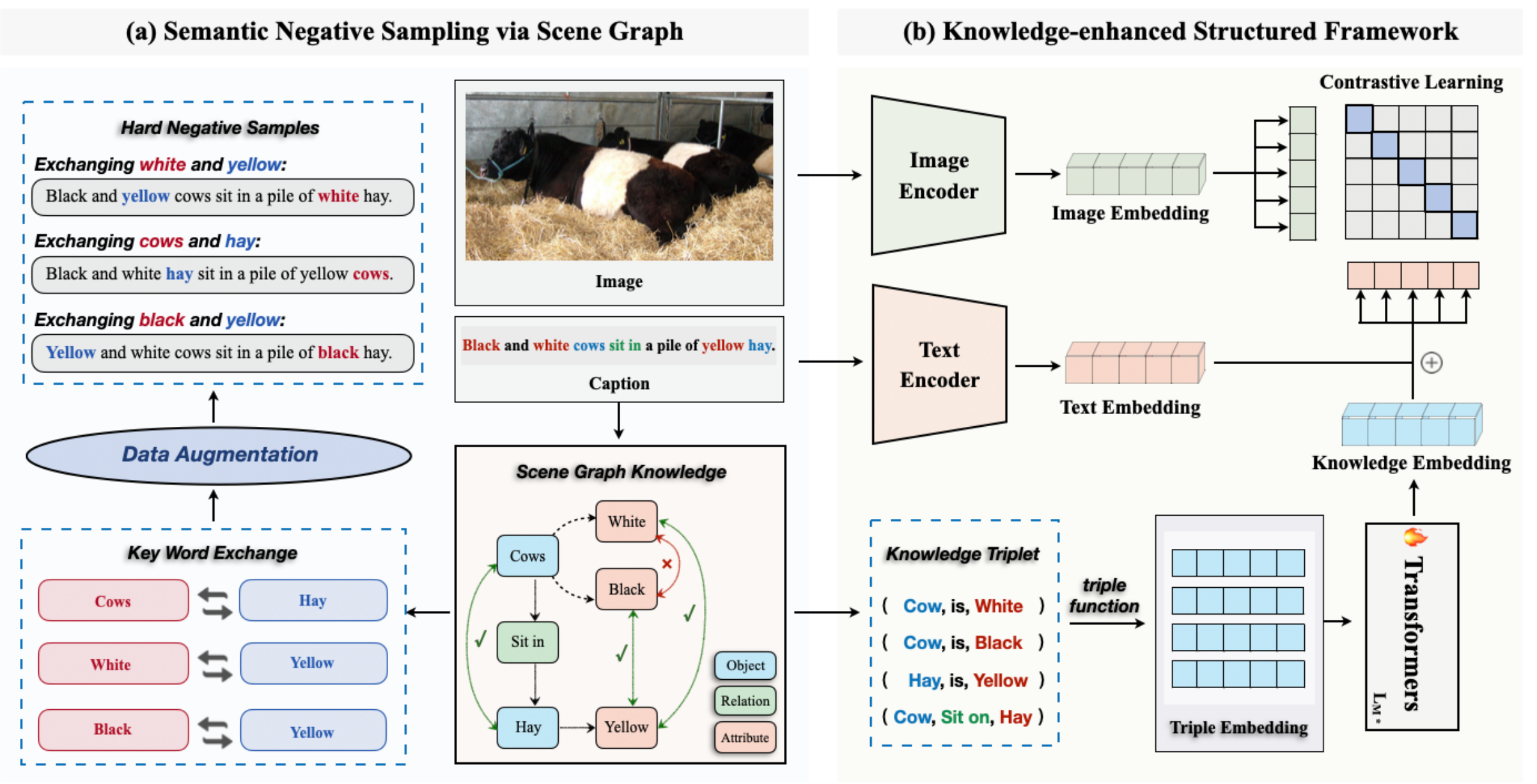}
  \caption{Overview of Structure-CLIP. (a) \emph{Semantic negative sampling via scene graph}: we extract a scene graph from the caption to help construct high-quality negative samples(left part).
  (b)\emph{Knowledge-Enhanced Encoder}:  \tjj{Knowledge embedding module and multiple} Transformers layers are used to model structured knowledge at the input level(right part).}
  \label{fig:model}
  \vspace{-10pt}
\end{figure*}


\section{Related Work}
\subsection{Vision Language Pretraining}
\tjj{\emph{Vision-Language Models} (VLMs)} aim to learn universal cross-modal representations, which are beneficial for achieving strong performance in downstream multi-modal tasks.
Depending on the multi-modal downstream task, different model architectures have been developed, including the dual-encoder architecture \cite{DBLP:conf/icml/RadfordKHRGASAM21,DBLP:conf/icml/JiaYXCPPLSLD21}, the fusion-encoder architecture \cite{DBLP:conf/emnlp/TanB19,DBLP:conf/nips/LiSGJXH21}, the encoder-decoder architecture \cite{DBLP:conf/icml/ChoLTB21,DBLP:conf/iclr/WangYYDT022,DBLP:journals/corr/abs-2209-06794}, and more recently, the unified transformer architecture \cite{DBLP:conf/icml/0001LXH22,DBLP:journals/corr/abs-2208-10442}.

The pre-training tasks have a great impact on what \tjj{VLMs} can learn from the data. 
\wen{There are mainly 4 types of tasks:}
\textit{(i)}  Cross-Modal Masked Language Modeling (MLM) \cite{DBLP:conf/icml/KimSK21,DBLP:journals/corr/abs-2003-13198,DBLP:conf/nips/LiSGJXH21,DBLP:journals/corr/abs-2205-01917};
\textit{(ii)}  Cross-Modal Masked Region Prediction (MRP) \cite{DBLP:conf/nips/LuBPL19,DBLP:conf/eccv/ChenLYK0G0020,DBLP:conf/cvpr/HuangZH0FF21};
\textit{(iii)} Image-Text Matching (ITM) \cite{DBLP:conf/aaai/LiDFGJ20,DBLP:conf/nips/LuBPL19,DBLP:conf/eccv/ChenLYK0G0020,DBLP:conf/cvpr/HuangZH0FF21};
\textit{(iv)} Cross-Modal Contrastive Learning (CMCL) \cite{DBLP:conf/icml/RadfordKHRGASAM21,DBLP:conf/icml/JiaYXCPPLSLD21,DBLP:conf/nips/LiSGJXH21,DBLP:journals/corr/abs-2103-06561,DBLP:conf/acl/LiGNXLL0020}.

\tjj{Recent researches mainly focus on the study of CMCL. Taking the CLIP model ~\cite{DBLP:conf/icml/RadfordKHRGASAM21} as an example, the model  learned sufficient general representations by comparing positive examples with  negative examples from all other samples in the dataset.}

\subsection{Structured Representation Learning}
Structured Representations denote the ability to match images and texts that have \tjj{identical word compositions}. 
Winoground \cite{DBLP:conf/cvpr/ThrushJBSWKR22} first presented a novel task and dataset for evaluating the ability of \tjj{VLMs}. The dataset comprises primarily 400 hand-crafted instances, where each instance includes two sentences with similar word compositions but distinct semantics, along with corresponding images.
The \tjj{evaluation} results of Winoground identified the dataset’s main challenges through a suite of experiments on related tasks (i.e., probing task, image retrieval task), suggesting that \hyf{the} main challenge in vision-language models may lie in fusing visual and textual representations, rather than in the understanding of compositional language.

Due to the limited quantity of Winoground test data, it is \hyf{challenging} to draw dependable experimental results \tjj{on \hyf{the} ability of structure representations}. \tjj{Recently,} NegCLIP ~\cite{DBLP:journals/corr/abs-2210-01936} provided a large-scale test bed to evaluate structured representations of VLMs. Additionally, NegCLIP also proposes a negative sampling method to enhance structured representations.

\subsection{Scene Graph Generation}
\tjj{A} scene graph is a \tjj{type of structured knowledge}, which describes the most essential parts of a multi-modal sample, through modeling objects, attributes of objects, and relations between objects and subjects. Generally, \emph{Scene Graph Generation}(SGG) models consist of three main modules: proposal generation localizing the bounding box of objects, object classification labeling the detected objects, and relationship prediction predicting the relations between pairwise objects.
Some existing works ~\cite{DBLP:conf/cvpr/XuZCF17,DBLP:conf/eccv/YangLLBP18,DBLP:conf/cvpr/ZellersYTC18} applied RNNs and GCNs to propagate image contexts in order to achieve better utilizing the contexts for object and relationship prediction. VCTree ~\cite{DBLP:conf/cvpr/TangZWLL19} captured local and global visual contexts by exploiting dynamic tree structures. \citet{DBLP:conf/cvpr/GuZL0CL19} and \citet{DBLP:conf/cvpr/ChenYCL19} integrated external knowledge into SGG models to address the bias of noisy annotations. 

As a beneficial prior knowledge describing the detailed semantics of images and captions, scene graphs have helped achieve excellent performance in several vision-language tasks.
Such as image captioning ~\cite{DBLP:conf/cvpr/YangTZC19}, image retrieval ~\cite{DBLP:conf/cvpr/0011MZJLSM19}, visual question answering ~\cite{DBLP:conf/bmvc/ZhangCX19,DBLP:journals/corr/abs-2205-11501}, multi-modal sentiment classifications ~\cite{DBLP:journals/corr/abs-2208-09417}, image generation ~\cite{DBLP:conf/cvpr/JohnsonGF18} and vision-language pretraining \cite{yu2021ernie}.

\section{Methodology}

The overview of Structure-CLIP is illustrated in Fig. \ref{fig:model}.
Firstly, our approach leverages the scene graph to enhance fine-grained structured representations by generating semantic negative samples with \tjj{identical word compositions} but differing detailed semantics(\emph{left part of Fig. \ref{fig:model}}). 
Secondly, we propose a Knowledge-Enhanced Encoder that utilizes the scene graph as an input to integrate structured knowledge into the structured representations (\emph{right part of Fig. \ref{fig:model}}).
We will introduce semantic negative sampling via the scene graph in Section 3.1 and present the Knowledge-Enhanced Encoder in Section 3.2.

\subsection{Semantic Negative Sampling via Scene Graph}
\citet{DBLP:conf/bmvc/FaghriFKF18} proposed 
a negative sampling method that involves 
constructing negative examples to enhance the representations by comparing them with positive samples. 
Our objective is to construct samples with \tjj{similar general representations} but differing detailed semantics, thereby encouraging the model to focus on learning structured representations.

\subsubsection{Scene Graph Generation.}
Detailed semantics, including objects, attributes of objects, and relationships between objects, are essential to the understanding of visual scenes.  And they are critical to cross-modal learning, which aims to enhance the joint representation of vision and language.
In our framework, the Scene Graph Parser provided by \cite{wu2019unified} is adopted to parse texts to scene graphs. Given the text sentence $\mathbf{w}$,  we parse it into a scene graph~\cite{DBLP:conf/cvpr/JohnsonKSLSBL15}, which denotes as  $G(\mathbf{w})=<O(\mathbf{w}),E(\mathbf{w}),K(\mathbf{w})>$, where $O(\mathbf{w})$ is the set of objects mentioned in $\mathbf{w}$, 
\wen{$R(\mathbf{w})$ is the set of relationship nodes, and $E(\mathbf{w})\subseteq O(\mathbf{w})\times R(\mathbf{w})\times O(\mathbf{w})$ is the set of hyper-edges representing actual relationships between objects.}
$K(\mathbf{w})\subseteq O(\mathbf{w})\times A(\mathbf{w})$ is the set of attribute pairs, where  $A(\mathbf{w})$ is the set of attribute nodes associated with objects. 


\wen{As shown in Figure \ref{fig:model}, we generate the scene graph based on the original caption. Using the caption ``Black and white cows sit in a pile of yellow hay"  in \tjj{Fig.} \ref{fig:model} as an example, in the generated scene graph,  }
the objects, such as ``cows" and ``hay" are the fundamental elements. The associated attributes, such as ``white" and ``yellow" characterize the color or other attributes of objects. Relations such as ``sit in" represent the spatial connections between objects. 


\subsubsection{Choice of Semantic Negative Samples.}
Contrastive learning aims to learn effective representations by pulling semantically close neighbors together and pushing apart non-neighbors. 
Our objective is to construct semantic negative samples with similar composition but different detailed semantics. Therefore, the quality of negative samples plays a vital role in structured representation learning.

A multi-modal dataset usually consists of N image-text pairs, where image and text are denoted as $I$ and $W$ with subscripts, respectively. Given an image-text pair $(I_i, W_i)$ and a related scene graph $G(W_i)$ generated from $W_i$, a high-quality semantic negative sample $W_i^-$ is generated via
\begin{equation}
W_i^-=F(W_i, G(W_i))\,,
\end{equation}
where $F$ is the proposed sampling function, $W_i^-$ denotes the high-quality semantic negative sample.
Specifically,  
for triples $(object, relation, subject)$ in the scene graph, $W_i^-$ is generated via
\begin{equation}
W_i^-=Swap((O_1, R, O_2)) = (O_2, R, O_1)\,,
\end{equation}
where  $Swap$ is the function to exchange object and subject in the sentence, $O_1, R, O_2$ denote the object, relation and subject. For attribute pairs 
\wen{$(A1, O1)$ and $(A2, O2)$}
 in the scene graph, $W_i^-$ is generated via

\begin{equation}
\small
    Swap((A_1O_1),(A_2O_2))=\left\{\begin{array}{l}
    (A_2O_1),(A_1O_2) \quad \text{ if }~O_1 \neq O_2,  \\
    pass \quad \text{ if }~O_1 = O_2,  \\ 
    \end{array}\right.
    \vspace{-2pt}
\end{equation}

Overall, we leverage scene graph guidance to construct high-quality semantic negative samples, instead of randomly swapping word positions. Our semantic negatives maintain the same sentence composition while altering detailed semantics. As a result, our model can more effectively learn structured representations of detailed semantics.


\subsubsection{Contrastive Learning Objective.}

Our contrastive learning objective is to learn sufficient representations by pulling image $I_i$ and origin caption $W_i$ together and pushing apart image $I_i$ and negative sample $W_i^-$.
Specifically, we introduce a multi-modal contrastive learning module with the loss function:
\begin{equation}
    \mathcal{L}_{hinge} = max(0, \gamma-d+d^\prime),
\end{equation} 
where $\gamma$ is the margin hyper-parameter, $d$ denotes the distance between image $I_i$ and origin caption $W_i$ and $d^\prime$ denotes the distance between image $I_i$ and origin caption $W_i^-$.
The contrastive learning objective is introduced to improve the performance of structured representations.
 Meanwhile, in order to maintain 
 \wen{the general representation ability}
 of the model, we combine the original 
 mini-batch 
 \wen{image-text}
 contrastive learning 
 \wen{loss} 
 and the proposed 
 \wen{loss}
 for joint training. 

 \wen{The original image-text contrastive learning loss $\mathcal{L}_{ITCL}$ contains an image-to-text constrastive loss $\mathcal{L}_{i2t}$ and a text-to-image contrastive loss $\mathcal{L}_{t2i}$ that
 \begin{equation} \label{eq.allloss}
  \mathcal{L}_{ITCL} = (\mathcal{L}_{i2t} + \mathcal{L}_{t2i})/2,
\end{equation}
 } 

The image-to-text contrastive loss $\mathcal{L}_{i2t}$ is formulated as
\begin{equation}
        \mathcal{L}_{i2t} = - \log \frac{\exp((\tilde{v}_i, e_{text_{i}}) / \tau)}{\sum^N_{k=1} \exp((\tilde{v}_i, e_{text_{k}}) / \tau)},
\end{equation}
where $\tau$ is the temperature hyper-parameter. Similarly, the text-to-image contrastive loss $\mathcal{L}_{t2i}$ is
\begin{equation}
        \mathcal{L}_{t2i} = - \log \frac{\exp((e_{text_{i}}, \tilde{v}_i) / \tau)}{\sum^N_{k=1} \exp((e_{text_{i}}, \tilde{v}_k) / \tau)},
\end{equation}
\wen{Thus the }
final loss, which \hyf{combines} the hinge loss and InfoNCE loss, is 
\begin{equation}
  \mathcal{L}_{final} \wen{=} \mathcal{L}_{hinge} + \mathcal{L}_{ITCL}. 
\end{equation}

\begin{table*}[!htbp]
\centering
\small
\setlength\tabcolsep{8pt}
\renewcommand\arraystretch{1}
\caption{Results ~($\%$) comparison between our method and other baselines on the VG-Relation, VG-Attribution and MSCOCO datasets. The matching scores are obtained via semantics similarities between image embeddings and text embeddings in multi-modal models and Maximum Likelihood Probability in large language models, respectively.}
\begin{tabular}{cccccccc}
\toprule
\multirow{2}{*}{\textbf{Domains}} &
\multirow{2}{*}{\textbf{Models}} & \multirow{2}{*}{\textbf{Params}} & \multicolumn{2}{c}{\textbf{Visual Gnome}} & & \multicolumn{2}{c}{\textbf{MSCOCO}}   \\
\cline{4-5}
\cline{7-8} 
& & & \small{Attribute} &  \small{Relation} & & \small{IR-R@1} & \small{TR-R@1} \\

\midrule
\multirow{1}{*}{-}  & Random Chance & - & 50.00 & 50.0 & & 0.02 & 0.1  \\
\midrule

\multirow{6}{*}{Multi-modal Models}  & VILT (VIT-B/32) & 87 M & 20.3 & 39.5 & & 37.3 & 53.4 \\
  & FLAVA & 241 M & 58.1 & 28.0 & & 38.5 & 43.5 \\
  
  & CLIP-Base (ViT-B/32) & 151 M & 60.1 & 59.8 & & 30.4 & 50.1   \\
  & CLIP-Large (ViT-L/14) & 427M & 61.1 & 61.5 & & 36.5 & 56.3  \\
  & Neg-CLIP & 151 M & 71.0 & 81.0  & & 41.0 & 56.0  \\
\midrule

\multirow{3}{*}{Large Language Models}  & BART & 300 M & 73.6 & 81.1  & & - & -  \\
  & FLAN-T5 & 11 B & 76.5 & 84.4 & & - & - \\
  & OPT & 175 B & 79.8 & 84.7 & & - & - \\
\midrule
\multirow{2}{*}{Ours}  & Sturcture-CLIP-Base &220 M & 82.3 & 84.7 & & 41.2 & 55.6 \\
  & Structure-CLIP-Large & 496 M & \textbf{83.5 } & \textbf{85.1} & & \textbf{48.9} & \textbf{58.2} \\
\bottomrule
\end{tabular}
\label{tab:main-results}
\end{table*}

\subsection{Knowledge-Enhanced Encoder}
In this section, we propose a Knowledge-Enhanced Encoder, which utilizes scene graphs as the textual input to enhance the structured representations.
To begin with, we use the following function to encode image $I_i$ and text $W_i$:
\begin{equation}
        \tilde{v} = CLIP_{vis}(I_i),     
\end{equation}
\begin{equation}
        \tilde{z} = CLIP_{text}(W_i),     
\end{equation}
where $CLIP_{vis}$ and $CLIP_{text}$ denote the visual encoder and text encoder of the CLIP model, respectively.

However, the CLIP model processes text input in a word-bag manner, which ignores the detailed semantics of the text. In contrast, incorporating a scene graph captures crucial structural information from the sentence, thereby enabling the model to gain deeper insights into the fine-grained semantics of the text.

Therefore, \hyf{the} \tjj{Knowledge-Enhanced Encoder} explicitly models the detailed knowledge as \tjj{model} input, i.e., objects, attributes of objects, and relations between paired objects.
Specifically, we make a unified input specification for two structured knowledge: pairs and triples. We add the relationship conjunction “is” to the pair to unify the representations. For example, The pair $ (white, cow) $ will be treated as the triple $ (cow, is, white) $ in this manner.
In this manner, a set of triples $\mathcal{T}_{in}=\{(h_i, r_i, t_i) | i \in [1,k]\}$ are obtained, \tjj{where $(h_i, r_i, t_i)$ represent \hyf{the} head entity, relation entity and tail entity respectively.}
For each triple $(h_i, r_i, t_i)$ in $\mathcal{T}_{in}$, we use \tjj{Tokenizer and Word Vocabulary Embeddings} from BERT \cite{DBLP:conf/naacl/DevlinCLT19} to obtain each \tjj{entity} embedding $w_h$, $w_r$, $w_t$:
\begin{equation}
w_{x} = WordEmb(x), x \in [h, r, t],
\end{equation}
In order to get the triple embedding with each entity embedding, we use the following encoding function:
\begin{equation}
\begin{aligned}
e_{triple_{i}} = ENC_{triple}(h_i, r_i, t_i) = w_{h,i} + w_{r,i} - w_{t,i},
\end{aligned}
\end{equation}
where $ENC_{triple}(.)$ is the triple encoding function.
With this triple encoder, our method can better solve the problem that the order of the head and tail entities is reversed, a detailed analysis is illustrated in Sec. 4.4.3.

In this way, K triples can be processed into K semantic embeddings. Then we input $e_{triple}$ to \tjj{multiple} Transformer layers to get the final representations. 
\begin{equation}
e_{KE} = TRMs([e_{triple_{1}}, ...,e_{triple_{K}}]),
\end{equation}
The \tjj{Knowledge-Enhanced Encoder} enables us to extract \tjj{sufficient structured knowledge} from all input triples, which can be utilized as effective structured knowledge to improve the performance of structured representations.

Thus, the Knowledge-Enhanced Encoder can be utilized to obtain text knowledge embedding\hyf{s}. However, relying solely on structured knowledge may result in a loss of representing general semantics. Therefore, \hyf{we} integrate both text embedding\hyf{s} and structured knowledge embedding\hyf{s} :
\begin{equation}
\begin{aligned}
    e_{text} 
  & = \tilde{z} + \lambda e_{KE}  \\
  & = CLIP_{text}(W_i) + \lambda \cdot TRMs([e_{triple_{*}}]),
\end{aligned}
\end{equation}
where $\lambda$ is a hyper-parameter, $\tilde{z}$ and $e_{KE}$ denote the origin text embedding and the structural knowledge embedding.

Our textual representations contain both the word information carried by the whole sentence and the structured knowledge composed of the detailed semantics in the sentence. Similarly, we used the same loss strategy illustrated in Eq. ~\ref{eq.allloss} in the training process.

\begin{table*}[htbp]
    \caption{Results ~($\%$) of ablation study on VG-Relation and VG-Attribution datasets to analyze different components. Results show that each component greatly improves the ability of structured representation. 
    }
    \small
    \centering
    \begin{tabular}{lccccccc}
    \toprule
    {\bf Methods}  & {\bf Finetune} & {\bf Negatives}  & {\bf KEE} &{\bf VG-Attribution}  & {\bf VG-Relation}  \\
    \toprule
     CLIP & \ding{55} & \ding{55} & \ding{55} & 60.1 & 59.8 \\
    CLIP (fine-tune) & MSCOCO (\emph{ours}) & \ding{55}  & \ding{55}   & 64.0 & 66.5 \\
    Neg-CLIP & MSCOCO (\emph{full}) & Random & \ding{55} &  71.0 & 81.0 \\
    \midrule
    w/ \{Random Change\} & MSCOCO (\emph{ours}) & Random  & \ding{55} & 73.9 & 77.7 \\
    
    w/ \{Semantic Negative\} &  MSCOCO (\emph{ours}) & Semantic   & \ding{55} & 77.8 & 79.0 \\
    
    w/ \{Transformer\} & MSCOCO (\emph{ours}) & \ding{55}  & \ding{51} & 65.7 & 68.8 \\
    {\textbf{Structure-CLIP}}~{\textbf{(Ours)}} & MSCOCO (\emph{ours}) & Semantic & \ding{51} &  \textbf{82.3} \small ($\uparrow 11.3$) & \textbf{84.7} \small ($\uparrow 3.7 $)\\
    \bottomrule
    \end{tabular}
    \label{tab:Ablation}
\end{table*}

\begin{table*}[htbp]
    \caption{Ablation study of different hyperparameters and embedding methods.}
    \small
    \centering
    \begin{tabular}{lccccc}
    \toprule
    {\bf Types} &
    {\bf KEE Layers} &
    {\bf Fusion Weight ($\lambda$)} &
    
    {\bf Embedding Fusion ($ENC_{triple}$)}  & {\bf VG-Attribution}  & {\bf VG-Relation}  \\
    \toprule
    \multirow{4}{*}{Layers }
    & 1 layer & 0.2 &  head + relation - tail & 82.1 & 82.9 \\
    & 2 layers  & 0.2 &  head + relation - tail & 82.2 & 83.3 \\
    & 6 layers & 0.2 &  head + relation - tail & \textbf{82.3} & \textbf{84.7} \\
    & 12 layers & 0.2 &  head + relation - tail &  81.9 & 83.2  \\
    \midrule
    \multirow{5}{*}{Weight }
    & 6 layers & 0.0 &  head + relation - tail & 77.8 & 79.0 \\
    & 6 layers  & 0.01 &  head + relation - tail & \textbf{82.7} & 83.5 \\
    & 6 layers & 0.2 &  head + relation - tail & 82.3 & \textbf{84.7} \\
    & 6 layers & 1.0 &  head + relation - tail & 82.3 & 83.8 \\
    \midrule
    \multirow{4}{*}{Embedding }
    & 6 layers & 0.2 &  Concat & 81.1 & 83.3 \\
    & 6 layers & 0.2 & head + relation + tail & 81.9 & 83.3 \\
    & 6 layers & 0.2 &  head + relation - tail & \textbf{82.3} & \textbf{84.7} \\
    
    \bottomrule
    \end{tabular}
    \label{tab:abaltions}
    \vspace{-10pt}
\end{table*}

\section{Experiments}
\subsection{Datasets}
\subsubsection{Pretraining Datasets.}
High-quality image-text alignment data is a critical aspect of training models.
We adopt the widely-used cross-modal text-image retrieval dataset, MSCOCO~\cite{DBLP:conf/eccv/LinMBHPRDZ14}. Consistent with prior work~\cite{DBLP:conf/icml/0001LXH22}, we utilize the Karpathy~\cite{DBLP:journals/pami/KarpathyF17} split for training and evaluation.
In our experiment, pre-training is conducted by filtering approximately 100k image-text pairs that involve multiple objects, attributes, and relationships. Subsequently, the models are evaluated on test splits, encompassing 5k images.  
We report Recall@1 on image-to-text retrieval(\textbf{IR}) and text-to-image retrieval(\textbf{TR}) to measure the ability of general representations.

\subsubsection{Downstream Datasets.}
Two novel datasets \cite{DBLP:journals/corr/abs-2210-01936} are used to evaluate the structured representation performance of different models,  where each test case consists of an image with matched captions and swapped mismatched captions. The model is tasked with distinguishing between aligned and unaligned captions based on the corresponding image.
\begin{itemize}[leftmargin=*]
\item \textbf{Visual Genome Relation (VG-Relation).}
As given an image and a caption containing a relationship triple, we evaluate the model's ability to select the caption where the relation is aligned with the image. Specifically, we expect the model to distinguish between ``$X$ relation $Y$" and ``$Y$ relation $X$" with a certain image ~(e.g., ``an \textbf{astronaut} is riding a \textbf{horse}" 
\wen{v.s.}
``a \textbf{horse} is riding an \textbf{astronaut}" with \hyf{the} image in Fig. ~\ref{fig:case}(a)).
\item \textbf{Visual Genome Attribution (VG-Attribution).}
Given the form ``$A_1$ $O_1$ and $A_2$ $O_2$" and ``$A_2$ $O_1$ and $A_1$ $O_2$", we evaluate the model's ability to accurately attribute the properties of objects. As shown in Fig. ~\ref{fig:case}(b), 
we expect the model to distinguish between the caption ``the \textbf{red} dress and the \textbf{blue} book" and the caption ``the \textbf{blue} dress and the \textbf{right} book" according to the image.
\end{itemize}

\subsection{Experimental Settings}

All of our experiments are performed on single NVIDIA A100 GPU with the Pytorch framework. 
We utilize a pre-trained Scene Graph Generator \cite{wu2019unified} to extract the Scene Graph Knowledge.

The structured \tjj{Knowledge-Enhanced Encoder} is implemented using a 6-layer Transformer architecture initialized with BERT-base \cite{DBLP:conf/naacl/DevlinCLT19}.

During the training stage, we initialize the model with a pre-trained CLIP model and train it on our dataset for 10 epochs using a batch size of 128. We use a mini-batch AdamW optimizer with a weight decay of 0.1. The learning rate is initialized as 2e-6. The knowledge weight $\lambda$ is 0.2.

\subsection{Overall Results}



\subsubsection{Structured Representation Tasks.} 

We compare our method with \emph{8} representative or SOTA methods, including multi-modal models and large language models.
As shown in 
\tjj{Table}
\ref{tab:main-results}, we note that our Structure-CLIP has achieved the SOTA performance over all baselines across VG-Relation and VG-Attribute datasets. 

Firstly, it is evident that NegCLIP outperforms the CLIP model in terms of structured representations, demonstrating that the aforementioned negative example sampling method can significantly enhance structured representations. Furthermore, by leveraging the guidance of Scene Graph Knowledge to improve the quality of constructing negative examples, Structure-CLIP achieves even further enhancement of structured representations. As a result, Structure-CLIP outperforms the existing multi-modal SOTA model (NegCLIP) by 12.5\% on VG-Attribution and 4.1\% on VG-Relation, respectively.

We also compare Structure-CLIP with existing Large Language models~(LLMs) which use Maximum Likelihood Probability for an image and a text
as the matching score. Our results demonstrate that as the model parameters of LLMs increase significantly, the structured representations also improve accordingly. However, Structure-CLIP still outperforms the OPT model by 3.7\% and 0.4\% respectively even though its parameters are less than 1\% of it. 
Our results indicate that increasing model parameters to improve structured representations is resource-intensive and yields suboptimal performance, as the model primarily learns general representations rather than structured representations during the training stage. In contrast, our proposed Structure-CLIP method can significantly enhance structured representations with only a minimal increase in model parameters and a small amount of training.

\subsubsection{General Representation Tasks.}


We evaluate the performances of Structure-CLIP on general representation tasks. Under the base model manner,  Structure-CLIP achieves comparable performances with NegCLIP on the MSCOCO dataset. In other words, while greatly improving the performance of structured representations, Structure-CLIP retains the ability of general representations. Furthermore, our results demonstrate that both adequate general representations and structured representations can be obtained simultaneously using Structure-CLIP, whereas previous models generate insufficient structured representations.
Under large model settings, our proposed method for domain fine-tuning significantly enhances both structured and general representations compared to the out-of-domain model.

\begin{figure*}[!htbp]
  \centering
  \includegraphics[width=1\linewidth]{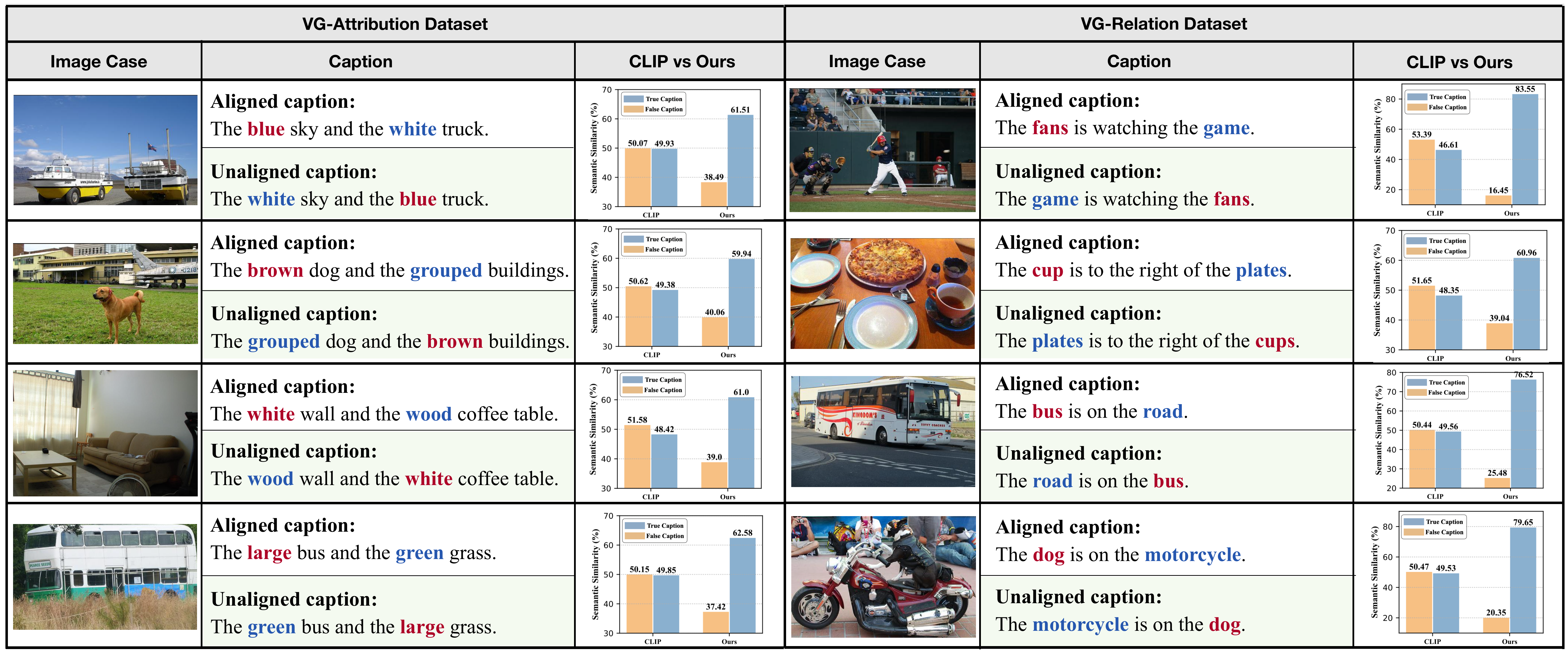}
  \caption{Predictions of different approaches. The words in red and blue are two exchanged words. We compare our structure-CLIP with CLIP to calculate CLIP scores (i.e., semantic similarity) between the image and captions. }
  \label{fig:prediction}
  \vspace{-10pt}
\end{figure*}

\subsection{Ablation Studies}
\subsubsection{Component Analysis.}
We perform an ablation study to assess multiple enhanced versions of the \tjj{CLIP-base} model on the VG-Relation and VG-Attribution datasets.
The results of each variant are presented in Table \ref{tab:Ablation}.

Firstly, our experimental results demonstrate a significant performance improvement when applying \emph{semantic negative} rather than \emph{random negative} sampling strategy~(\emph{Line 4 vs Line 5}). 
A notable increase of 3.9\%, 1.3\% on VG-Attribution and VG-Relation datasets
indicates that the proposed approach generates higher quality negative examples, resulting in superior structured representations. 

Incorporating structured knowledge as input via the proposed \tjj{Knowledge-Enhanced Encoder} yields only a  slight improvement (\emph{Line 2 vs Line 6}). These findings imply that in order to achieve adequate structured representations, the incorporation of negative example sampling is necessary. Therefore, the Knowledge-Enhanced Encoder achieves a significant enhancement after combining it with semantic negative sampling (\emph{Line 5 vs Line 7}).

\subsubsection{Hyperparameter Analysis.}
Based on the experiment results of Structure-CLIP presented in Table \ref{tab:abaltions}, we can draw the following conclusion:
\textit{(i)} As the number of knowledge transformer layers increases, the model's capacity to represent multi-modal structured representations improves. However, it is important to note that beyond a certain threshold, the available data may become insufficient to support the model's increased capacity, leading to potential over-fitting.
\textit{(ii)} The experimental results demonstrate that without structured knowledge integration, the performance of the model is unsatisfactory (\emph{Line 5}). Conversely, when structured knowledge is integrated, the variance in performance across different weights is minimal, indicating the effectiveness and intuitiveness of our method in enhancing structure representations.

\subsubsection{Triple embeddings.}
We explored three different ways of triple embedding \tjj{for} integrating triples. 
The \emph{concat} approach considers the order of input triple elements but fails to take into account the combination of head entities, \tjj{relation entities}, and tail entities. 
The \emph{head + relation + tail} methods incorporate the combination relationship among triples. However, they lack the ability to distinguish the order of triples. For example, \tjj{the final embeddings of two triples}, $(cow, is, white)$ and $(white, is, cow)$ are identical, which can not help the model to make a distinction.
Compared with these approaches, our triple embedding method takes into account both location and composition information. In this way, our Structure-CLIP model is better able to leverage structured knowledge within sentences to capture fine-grained semantic information and enhance multi-modal \tjj{structured} representations.

\subsection{Case Study}

The prediction \tjj{results} of the cases in VG-Relation and VG-Attribution are presented in Fig. \ref{fig:prediction}, which clearly illustrates that Structure-CLIP can successfully distinguish between \tjj{aligned} and \tjj{unaligned} captions as given \hyf{an} image in a very large margin.
However, the CLIP model encounters challenges in accurately determining the \tjj{semantic similarities between these captions and the given image}.
In particular, the CLIP model exhibits near-uniform semantic \tjj{similarities} when two attributes or objects are swapped, indicating a lack of capacity to capture structured semantics.
In contrast to the CLIP model, Structure-CLIP exhibits sensitivity to modifications in fine-grained semantics, indicating its ability in representing structured knowledge.
As an example, the caption ``the blue sky and the white truck" is used to evaluate the ability of Structure-CLIP to distinguish between \tjj{aligned} and \tjj{unaligned} captions when two attributes (i.e., blue and white) are exchanged. The results show that Structure-CLIP is able to make \tjj{a distinction} between \tjj{aligned} and \tjj{unaligned} caption with a margin of 25.16\%, which further verifies the effectiveness of the proposed method in enhancing multi-modal structured representations.

\section{Conclusion}
In this paper, we propose Structure-CLIP aiming to integrate Scene Graph Knowledge to enhance multi-modal structured  representations. Firstly, we use scene graphs to guide the \tjj{construction} of semantic negative examples. Additionally, we introduce a \tjj{Knowledge-Enhanced Encoder} that leverages Scene Graph Knowledge as input, thereby further enhancing the structured representations.
Our proposed Structure-CLIP outperforms all recent methods on pre-training tasks and downstream tasks, which illustrates that Structure-CLIP can effectively and robustly understand the detailed semantics in multi-modal scenarios.

\bibliography{aaai24}

\clearpage
\appendix

\section{Appendix}

\subsection{Dataset Statistics}
\subsubsection{\textbf{Pretraining Datasets.}}
High-quality image-text alignment data is a critical aspect of training the model.
We zoom into the standard cross-modal text-image retrieval dataset, namely MSCOCO\cite{DBLP:conf/eccv/LinMBHPRDZ14}. Following prior work\cite{DBLP:conf/icml/0001LXH22}, we use the Karpathy\cite{DBLP:journals/pami/KarpathyF17} split for this dataset. 
Given the significance of fine-grained semantics in our context, pre-training was performed by filtering nearly 100k image-text pairs containing attributes and relationships. Subsequently, the models were evaluated on test splits, encompassing 5k images.  
We report Recall@1, Recall@5 and Recall@10 for both image-to-text retrieval and text-to-image retrieval.

\subsubsection{\textbf{Downstream Datasets.}}
we evaluate two novel datasets\cite{DBLP:journals/corr/abs-2210-01936} for probing relation and attribution understanding. 
Each test case is thus made of an image (e.g., the image of an astronaut riding a horse) and a matched caption (e.g., “an astronaut riding a horse”) with the correct order, but swapped, mismatched caption (e.g., “a horse riding an astronaut”). 
For every test case included in these datasets, we quantify the performance of each model in identifying the correct caption from the two alternatives.

\subsection{Supplementary for Baselines}
We adopt three types of baselines. The details of each baseline are listed below:

\subsubsection{\textbf{Random Chance.}}
For every test case included in these datasets, we quantify the performance of each model in identifying the correct caption from the two alternatives, i.e. random chance level performance here is 50\%.
\subsubsection{\textbf{Multi-modal Models.}}
\begin{itemize}
    \item \textit{VILT} ~\cite{DBLP:conf/icml/KimSK21} greatly simplifies the processing of visual input in the same convolution-free way we process text input. We report results for the ``VIT-B/32" variants of VILT.
    \item \textit{FLAVA} ~\cite{DBLP:conf/cvpr/SinghHGCGRK22}  learns strong representations through joint pretraining on both unimodal and multi-modal data while encompassing cross-modal ``alignment" objectives and multi-modal ``fusion" objectives. We use the model released at the Huggingface Transformers library. We follow the tutorial shared by the authors and use the ``FLAVA-full" variant.
    \item \textit{BLIP} ~\cite{DBLP:conf/icml/0001LXH22} pre-trains a multi-modal mixture of the encoder-decoder model using a dataset bootstrapped from large-scale noisy image-text pairs by injecting diverse synthetic captions and removing noisy captions. We report results for the ``Base" variants of BLIP.
    \item \textit{CLIP} ~\cite{DBLP:conf/icml/RadfordKHRGASAM21} learns the alignment between text and images by contrast learning. We use the ``ViT-B/32" variant of CLIP.
    \item \textit{Neg-CLIP} ~\cite{DBLP:journals/corr/abs-2210-01936} proposes a simple fix: mining of composition-aware hard negatives.
\end{itemize}
\subsubsection{\textbf{Large Language Models.}}
\begin{itemize}
    \item \textit{BART} ~\cite{yuan2021bartscore} is a baseline under a generation-based paradigm. We pass captions into a pure LLM to compute text-only GPTScore.
    \item \textit{FLAN-T5} ~\cite{chung2022scaling} has strong generalization performance so that a single model can perform well on more than 1800 NLP tasks.
    \item \textit{OPT} ~\cite{zhang2205opt} is a suite of decoder-only pre-trained transformers ranging from 125M to 175B parameters, which aim to fully and responsibly share with interested researchers.
\end{itemize}

\begin{table*}[htbp]
\centering
\setlength\tabcolsep{12pt}
\renewcommand\arraystretch{1}
\caption{Results ~($\%$) of fine-tuned image-text retrieval on MSCOCO datasets. Our Structure-CLIP performed comparably well with CLIP-finetune. }
\label{tab:finetuned-retrieval-table}
\begin{tabular}{llllllllllll}
\midrule
\multirow{3}*{\textbf{Methods}}& \multicolumn{6}{c}{MSCOCO} \\
\cline{2-7}
&  \multicolumn{3}{c}{Text} & \multicolumn{3}{c}{Image} \\
\cline{2-7}
& R@1 & R@5 & R@10 & R@1 & R@5 & R@10  \\
\midrule
CLIP & 30.4 & 55.9 & 66.9 & 50.1 & 75.0 & 83.5 \\
CLIP-finetune  & 40.9 & 67.7 & 78.1 & \textbf{57.3} & \textbf{80.4} & 87.3 \\
Neg-CLIP  & 41.0 & - & - & 56.0 & - & - \\
{\textbf{Structure-CLIP}}~{\textbf{(Ours)}} & \textbf{41.2} & \textbf{68.5} & \textbf{78.6} & 55.6 & 79.4 & \textbf{87.6}\\
\bottomrule
\end{tabular}
\end{table*}

\subsection{Semantic Negative vs Hard Negative }

\begin{figure}
  \centering
  \includegraphics[width = 1\linewidth]{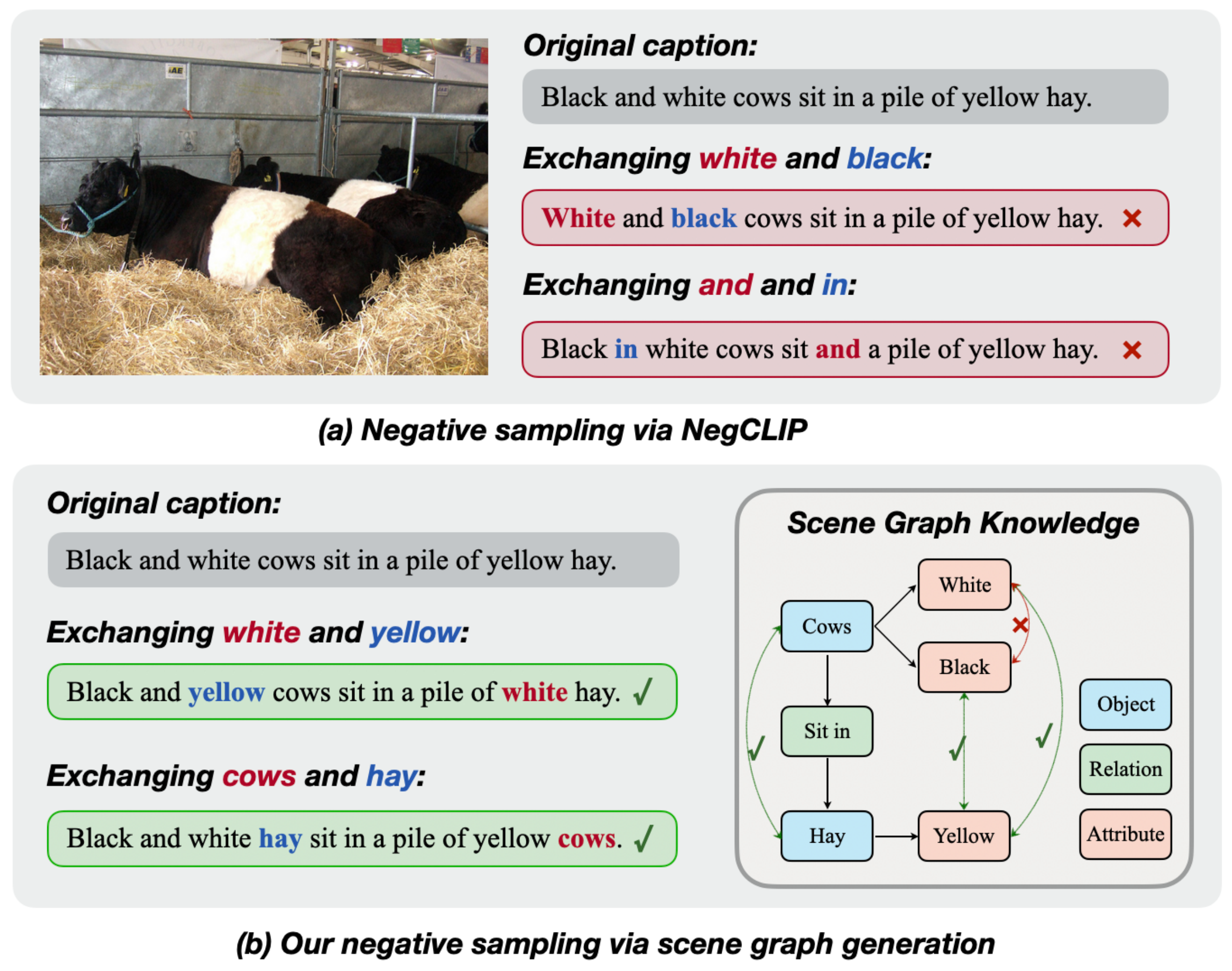}
  \caption{Our method is compared to NegCLIP in a negative sampling scenario. (a) Negative sampling in NegCLIP. We show two situations, one in which two attributes describing the same object are exchanged, and the other in which two less important words are exchanged (e.g., Prepositions, conjunctions). (b) Negative sampling via scene graph generation. From the caption, we get a scene graph from which we can capture the important pair suitable for the exchange.}
  \label{fig:sampling-problem}
\end{figure}

In this section, we will provide a detailed introduction to both \textbf{semantic negative} sampling and \textbf{hard negative} sampling methods. Additionally, the overview is illustrated in Fig.~\ref{fig:sampling-problem}.

The hard negative sampling method employed in NegCLIP obtains negative samples through random word swapping. However, this approach lacks semantic knowledge and could potentially introduce significant errors when generating negative examples. As depicted in Figure~\ref{fig:sampling-problem}, the negative sample generated by swapping the words "white" and "black" can inadvertently result in a positive sample. Moreover, swapping the words "in" and "and" renders the sentence semantically ambiguous and grammatically unclear. 

Consequently, the quality of the negative examples generated has a substantial effect on the final model's performance. Hence, we propose a semantic negative sampling method that incorporates structural knowledge when selecting words to swap.
Specifically, we first extract structured knowledge~(\textbf{Scene Graph Knowledge}) from text sentences,  which includes objects, attributes and relationships, as illustrated in the bottom right of Fig.~\ref{fig:sampling-problem}. And then we use scene graph knowledge to guide the construction of negative examples. For example, the word ``white" and ``yellow" are related to different objects, therefore we swap these two words to get semantic different object-attribute pairs. And we also swap objects (``cows") and subjects (``hay") related to the relationship word (``sit in"). As a result of it, our method can generate high-quality semantic negative samples.

\begin{figure}
  \centering
  \includegraphics[width = 1\linewidth]{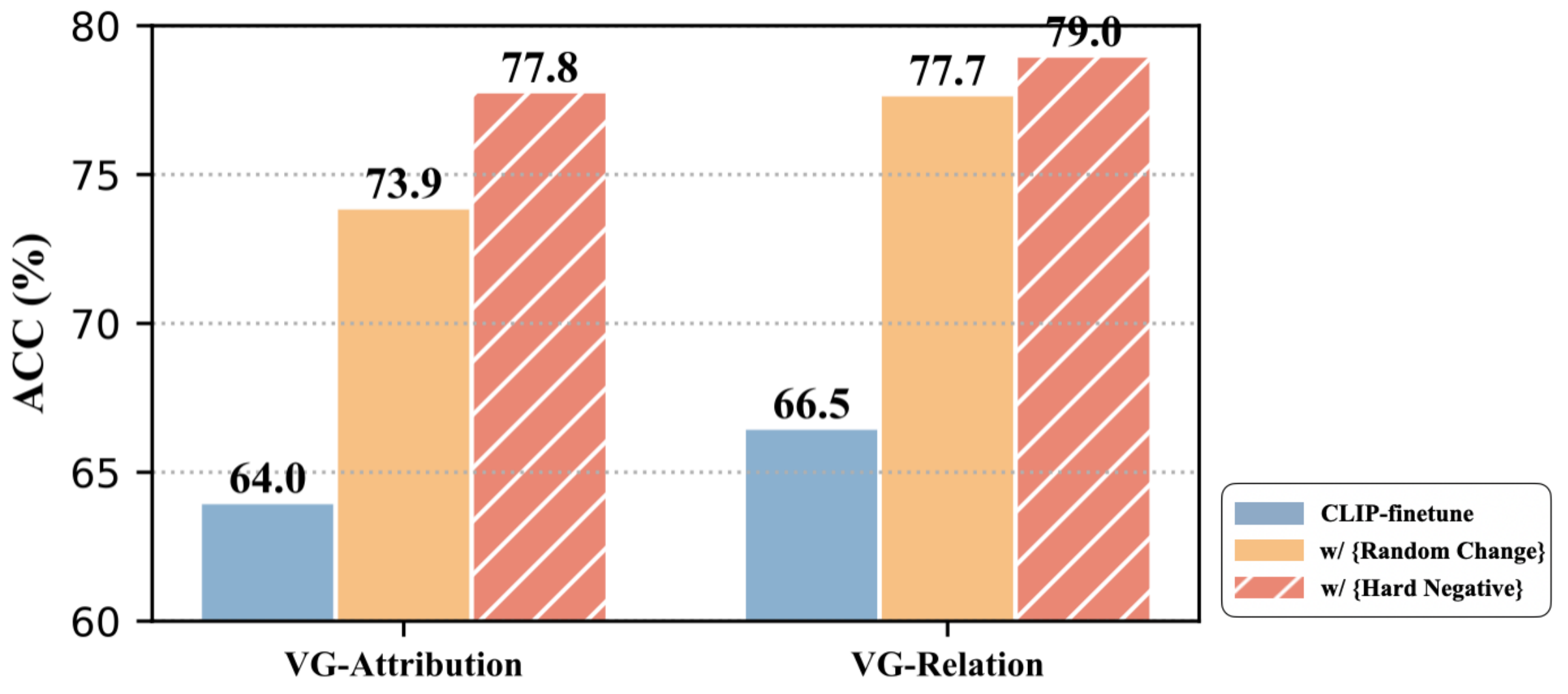}
  \caption{Ablation study of different negative samplings. Our hard negative sampling is more effective than others.}
  \vspace{-10pt}
  \label{fig:ablation-negative}
\end{figure}

We employ Figure~\ref{fig:ablation-negative} to visually demonstrate the effectiveness of our sampling method. Our results illustrate that fine-tuning without the use of any additional sampling method has minimal impact on structured representations. Furthermore, the structured representations significantly improve after introducing the strategy of random exchange (The outcome depicted in Figure~\ref{fig:ablation-negative} varies slightly from the result obtained in NegCLIP, as we implemented the hard negative method in our constructed dataset, as detailed in Section 4.1.).  Nevertheless, this approach overlooks the structured knowledge present in the sentence, thereby preventing a comprehensive exploration of the contrastive learning method's impact. In contrast, our proposed semantic negative method leverages high-quality negative examples to enhance structured representations, resulting in an improvement of 3.9\% on VG-Attribution and 1.3\% on VG-Relation.

\subsection{Supplementary for Experiments}
\subsubsection{Extending results on general representation tasks.}
we fine-tuned all models on a subset of the training split of the MSCOCO dataset, and then validated on the validation split. Since our proposed method requires the construction of negative samples based on either object in triples or adjectives in pairs, we also considered the special case of two adjectives modifying one noun when constructing negative samples. To address these issues, we used a subset of the training split of the MSCOCO dataset as our finetune data corpus.

Image-text retrieval consists of two sub-tasks: image-to-text retrieval and text-to-image retrieval. We evaluate our Structure-CLIP model on one retrieval benchmark dataset: MSCOCO, under fine-tuned settings. 
Results of the fine-tuned image-text retrieval are presented in Table \ref{tab:finetuned-retrieval-table}. Our comparisons included previous methods such as CLIP, CLIP-finetune, and Neg-CLIP. The results indicated that Structure-CLIP performed comparably well on the MSCOCO dataset. Notably, it outperformed CLIP-finetune on the Recall@10 of both text retrieval by $0.5$ points and image retrieval by $0.3$ points. These findings suggest that Structure-CLIP not only enhances the ability to capture detailed semantics in downstream tasks, but also improves the overall performance of multi-model pre-training.

\subsubsection{Extending method on other backbones.}
\begin{table}[htbp]
    \caption{Experiments on two different backbones. BLIP-based models exhibit similar performance trends as the CLIP-based models. The performance of Structure-BLIP surpasses the SOTA model NegCLIP of 19.3\% and 7.2\% on two datasets.}
    \small
    \centering
    \setlength\tabcolsep{4pt}
    \begin{tabular}{lcccc}
    \toprule
     {\bf BackBone} &
    {\bf Methods} 
    & {\bf VG-Attribution}  & {\bf VG-Relation}  \\
    \toprule
    \multirow{4}{*}{CLIP}
    & Base & 60.1 & 59.8 \\
    & Base-Finetune & 64.00 & 66.5 \\
    & Neg-CLIP & 71.0 & 81.0 \\
    & \textbf{Structure-CLIP} & \textbf{82.3} & \textbf{84.7} \\
    \midrule
    \multirow{3}{*}{BLIP}
    & Base & 74.3 & 50.4 \\
    & Base-Finetune & 75.2 & 62.1\\
    & \textbf{Structure-BLIP} & \textbf{90.3} & \textbf{88.2} \\
    \bottomrule
    \end{tabular}
    \label{tab:blipresults}
\end{table}

We conducted Structure-CLIP experiments on the CLIP model backbone and further verified the method's portability by conducting experiments on other backbones. Table~\ref{tab:blipresults} also shows the results of our experiments on the BLIP backbone.

Firstly, we observe that the BLIP-based models exhibit similar performance trends as the CLIP-based models. Specifically, the original BLIP model performs poorly on two structured representation downstream tasks, achieving only slightly better results than random chance. Fine-tuning the model with the previous general representation method leads to a slight improvement in its performance, whereas training with our proposed method results in a significant enhancement in structured representations. 
This demonstrates that our proposed method can effectively enhance structured representations and is transferable to a wide range of multi-modal models.

Moreover, by applying our proposed method to the superior base model BLIP, we achieve superior structured representations with the performance of 90.3\% and 88.2\% on VG-Attribution and VG-Relation, respectively, surpassing that of SOTA model NegCLIP of 19.3\% and 7.2\%.

\subsection{Future Work}
Knowledge Graphs (KGs) have been empirically validated to provide substantial benefits in several downstream applications. They serve as significant sources of knowledge supplementation and data augmentation for diverse tasks such as Question Answering \cite{DBLP:conf/semweb/0007CGPYC21,DBLP:conf/jist/0007HCGFP0Z22}, and Zero-shot Learning \cite{DBLP:journals/pieee/ChenGCPHZHC23,DBLP:conf/ijcai/ChenG0HPC21,chen2023duet,DBLP:conf/www/GengC0PYYJC21}.
We also note that there exist various in-KG tasks, such as 
Entity Alignment \cite{chen2022meaformer,chen2023rethinking} and Link Prediction \cite{zhang2023maco}.

Although the Scene Graph Knowledge in the text is helpful for fine-grained image-text matching, the knowledge is still limited to the fixed sentence itself. Therefore, we consider establishing the connection between the knowledge graph and the detailed semantics in the sentence, allowing external knowledge to guide the image and text matching so that the model can fully understand the connection between the detailed information in the sentence and the image.

Moreover, we believe that our proposed approach can be extended from the domain of multi-modal content understanding to the text-to-image generation task. We plan to investigate how to incorporate fine-grained semantic information into the image and leave this as a direction for future research.

\end{document}